\documentclass{article}

\usepackage{neurips_2021}

\usepackage[utf8]{inputenc} 
\usepackage[T1]{fontenc}    

\usepackage{url}            
\usepackage{booktabs}       
\usepackage{amsfonts}       
\usepackage{nicefrac}       
\usepackage{microtype}      
\usepackage{xcolor}         
\usepackage{graphicx}
\graphicspath{ {img} }
\usepackage{wrapfig}

\title{Brief Introduction to Contrastive Learning Pretext Tasks for Visual Representation}

\begin{document}

\maketitle

\begin{abstract}
To improve performance in visual feature representation from photos or videos for practical applications, we generally require large-scale human-annotated labeled data while training deep neural networks. However, the cost of gathering and annotating human-annotated labeled data is expensive. Given that there is a lot of unlabeled data in the actual world, it is possible to introduce self-defined pseudo labels as supervisions to prevent this issue. Self-supervised learning, specifically contrastive learning, is a subset of unsupervised learning methods that has grown popular in computer vision, natural language processing, and other domains. The purpose of contrastive learning is to embed augmented samples from the same sample near to each other while pushing away those that are not. In the following sections, we will introduce the regular formulation among different learnings. In the next sections, we will discuss the regular formulation of various learnings. Furthermore, we offer some strategies from contrastive learning that have recently been published and are focused on pretext tasks for visual representation.
\end{abstract}

\section{Introduction}

Large-scale dataset collection and annotation are time-consuming and costly. To avoid time-consuming and costly data annotations, a number of self-supervised learning methods have recently been developed to learn visual representations from massive unlabeled photos or videos that are not involved in human annotations. One frequent way of learning such visual representations is to propose a pretext task for the neural network to perform with. Here, we leverage contrastive learning to focus on the pretext task.

Consider Robert Epstein's experiment, in which the goal is to encourage participants to draw a detailed representation of a one-dollar bill (Figure 1). The image sketched for the dollar bill from memory is depicted in the figure on the left. While the dollar bill is presented, the correct figure is precisely drawn. As a result, the drawing produced by memory differs significantly from the drawing produced by the target presented (Epstein 2016). Regardless of how dissimilar these two pictures are, they share common representations such as Mr. Washington's figure, the one-dollar inscription, and others. Humans can comprehend that these two drawings depict the same target, one dollar. But what if we let the machine guess whether they are from the same image, which may require a representation based on a pair of positive sample pairs: a drawing and a dollar bill, and a pair of negative sample pairs: a random other drawing and a dollar bill. This is the concept of contrastive learning, which has lately been expanded to various algorithms.

\begin{figure}
  \centering
  \fbox{\includegraphics[scale=1]{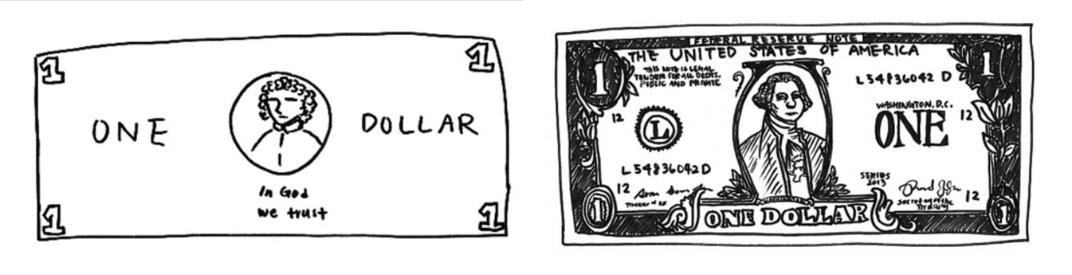}}
  \caption{Fig. Left: Drawing of a dollar bill from memory. Right: Drawing subsequently made with a dollar bill present. Image source: Epstein, 2016.}
\end{figure}

\section{Formulations Among Different Learning Paradigms}

The distinction between different learnings is primarily determined by training labels. There are four types of visual feature learning methods: (1) supervised learning, (2) semi-supervised learning, (3) weakly supervised learning, and (4) unsupervised learning (e.g. contrastive learning).

\subsection{Supervised Learning}
For supervised learning, the model is given a dataset $X\in(x_1,x_2,…,x_N )$. Such dataset is associated with manually annotated labels $Y_i$. The training loss function is defined as follows:

$$loss(D)=\min_{\theta} \frac{1}{N} \sum_{i=1}^N loss(X_i,Y_i )$$

where $D=\{X_i\}_{i=0}^N$  is the $N$ labeled training data. The advantage of training models with human-annotated labels is that they produce significant outcomes in a variety of computer vision applications (A. Krizhevsky 2012, R. Girshick 2014, D. Tran 2015, J. Long 2015). However, label annotation is frequently extremely expensive, demanding advanced professional skills and domain expertise. As a result, the other four learning algorithms are now more common than supervised learning for lowering labeling costs.

\subsection{Semi-supervised Learning}

The model is given a small labeled dataset X and a large unlabeled dataset Z for semi-supervised learning. This dataset is associated with manually annotated labels Y i. The following is the definition of the training loss function:

$$loss(D_1, D_2)=\min_{\theta} \frac{1}{N} \sum_{i=1}^N loss(X_i,Y_i )+\frac{1}{M}	\sum_{i=1}^{M}loss(Z_i, R(Z_i, X))$$

where $D_1=\{X_i\}_{i=0}^N$ is $N$ labeled training dataset, and $D_2=\{Z_i \}_{i=0}^M$ is $M$ unlabeled training dataset. $R(Z_i,X)$ is a function that represents the relationship between the unlabeled and labeled training datasets.

\subsection{Weakly Supervised Learning}

A dataset $X$ is associated with a collection of coarse-grained labels $C_i$ for weakly supervised learning. The training loss function for $X\in (x_1,x_2,…,x_i )$ is defined as follows:

$$loss(D)=\min_{\theta} \frac{1}{N} \sum_{i=1}^N loss(X_i,C_i )	$$

where $D_1=\{X_i\}_{i=0}^N$ denotes the training dataset. In a weakly supervised learning system, a fine-grained label is substantially more expensive than a coarse-grained label. Because of this fact, the advantage of weak supervision labels is that it is relatively easier to gather large-scale datasets. For example, picture features collected from websites utilizing the hashtag as coarse-grained labels were recently introduced (W. Li 2017, D. Mahajan and Y. Li 2018).

\subsection{Unsupervised Learning}

Unsupervised learning does not require human-annotated labels. Such techniques, like self-supervised learning, produce pseudo labels for grouping without any manually provided labels for the dataset involved. Another example is contrastive learning (We will introduce in the last section), which requires a huge amount of unlabeled data contrasted to certain identical labels in order to generate pseudo labels (e.g. this is not necessarily involved human annotations).

\subsubsection{Self-supervised Learning}
In self-supervised learning, a model is given a set of training data $X_i$ corresponding to its pseudo label $P_i$ that is generated autonomously for a pre-defined pretext task with no human annotations. Such labels $P_i$ can be constructed using image features  (D. Pathak 2016, Favaro 2016, R. Zhang 2016, S. Gidaris 2018) or other specified approaches (J. S´anchez 2013).

The training loss function is defined as:

$$loss(D)=\min_{\theta} \frac{1}{N} \sum_{i=1}^N loss(X_i,P_i )	$$

where $D=\{P_i\}_{i=0}^N$ is $N$ training dataset with pseudo labels $P_i$.

\section{Contrastive Learning}

\begin{figure}
  \centering
  \fbox{\includegraphics[scale=0.5]{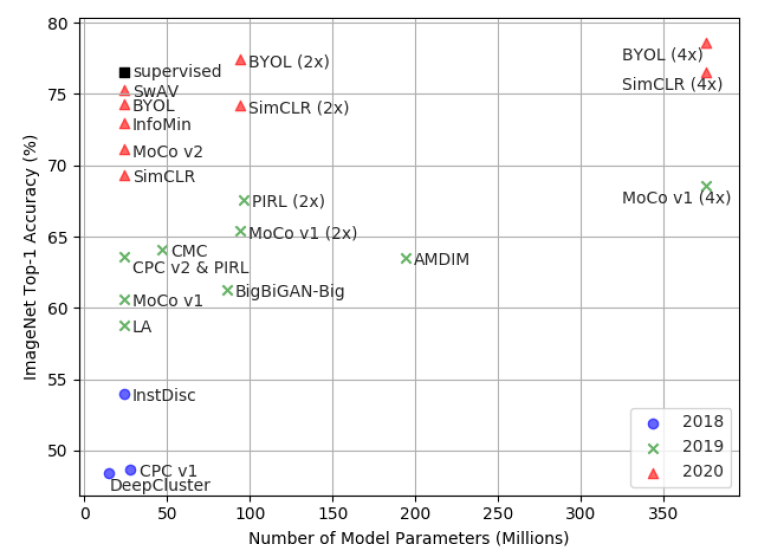}}
  \caption{Comparison of Self-supervised representation learning performance.}
\end{figure}

Machine learning algorithms can be classified statistically into generative and discriminative models. The generative method computes $p(X\|y)$ given the joint distribution $P(X,Y)$, $X$ as the input variables, and $Y$ as the labels, whereas the discriminative method computes $p(Y\|x)$. For a long time, most academics felt that developing models that understood the relationship between x was the only way to achieve representation learning.

Some recent contrastive learning studies (e.g., SimCLR, Deep InfoMax) demonstrate that discriminative models have the potential for representation learning (Chen, Kornblith et al. 2020). The most prevalent method in contrastive learning is Noise Contrastive Estimation (NCE), which is structured as:

$$L=\mathbf{E}_{x, x^+, x^-}[-log(\frac{e^{f(x)^T f(x^+)}}{e^{f(x)^T f(x^+)}+e^{f(x)^T f(x^-)}}]$$

where $x^+$ is identical to $x$, $x^-$ is not, and $f$ is the encoder. In actuality, the similarity estimation and encoder may differ from instance to case.

*Figure 2* shows the performance of self-supervised representation learning on ImageNet top-1 accuracy, using the linear classification technique. The capacity of self-supervised learning to extract features is fast nearing that of the supervised technique (ResNet50). All of the models listed above, with the exception of BigBiGAN, are contrastive self-supervised learning approaches (Liu, Zhang et al. 2020).

\subsection{Pretext Tasks}

\subsubsection{Instant Discrimination}

\begin{wrapfigure}{r}{0.5\textwidth}

	\centering
	\fbox{\includegraphics[width=0.48\textwidth]{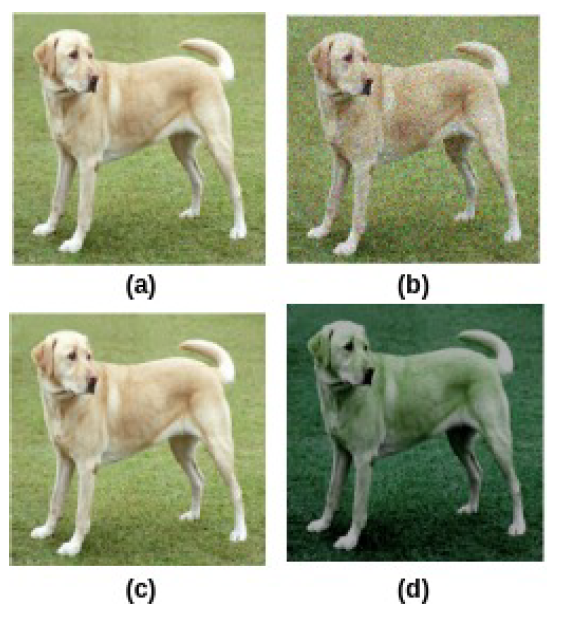}}

	\caption[width=0.2\linewidth]{Color transformation as pretext task.}

\end{wrapfigure}

Color transformation produces basic transformations of color levels in an image, such as gaussian blurring, Gaussian noise, color distortion, grayscale conversion, and so on (Figure 3). One example of how a model learns to distinguish similar images with different hues (Chen, Kornblith et al. 2020). The authors of this SimCLR study emphasize the relevance of a positive sample by offering data augmentation in ten different forms (Figure 4). This data augmentation uses multiple perspectives to augment the positive pairs. SimCLR chooses $N$, 8196 as the batch size.

The pairwise contrastive loss is defined as:

$$l_{i,j}=-log \frac{exp(sim(\hat{x_i}, \hat{x_j})/\tau)}{\sum_{k=1}^{2N}1_{[k\neq i]} exp(sim(\hat{x_i}, \hat{x_k})/\tau)} $$

where $\hat{x_j} (j=1,2,…,2N)$ is the augmentation of N samples. $\hat{x_i}$ and $\hat{x_j}$ is a pair of positive examples derived from one original sample. The rest $2(N-1)$ are treated as negative ones.

The summed-up loss:

$$L=\frac{1}{2N} \sum_{k=1}^N [l_{2i-1,2i}+l_{2i,2i-1}]$$

\begin{figure}
  \centering
  \fbox{\includegraphics[scale=0.35]{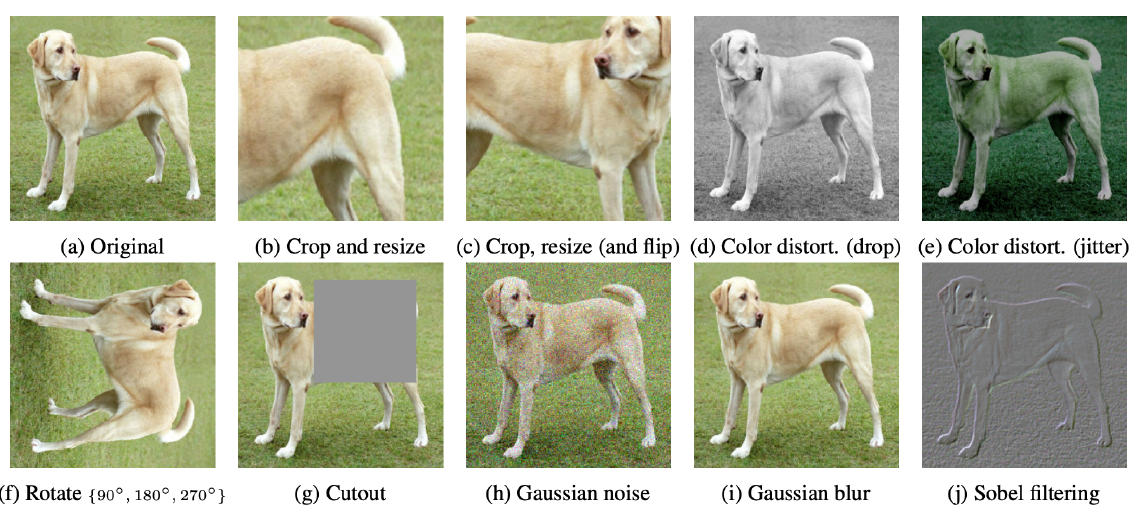}}
  \caption{Ten different views adopted by SimCLR.}
\end{figure}

\subsubsection{Global-Local Contrast}

The global-local contrast, also known as the context-instance contrast, is particularly concerned with assessing the relationship between sample local features and global feature representation. The purpose of this type of global-local comparison learning is to learn the local feature representation that should be linked with global content. We will introduce three different implementations of this learning in the following content: (1) jigsaw puzzle, (2) frame order-based learning, and (3) future prediction task.

\paragraph{Jigsaw Puzzle}
Figure 5 depicts the original image (a) and the reshuffled image (b), which is the positive sample to the (a). Solving Jigsaw puzzles is a challenge in learning local features from a global (e.g. original image) using an unsupervised method. The general process of this algorithm is to estimate the encoder to determine the right relative location of the reshuffled patches from a picture. The original image is fixed as an anchor in contrastive learning, and the augmentation of the original image is generated by reshuffled patches as positive samples. The remaining images/content were then viewed as negative samples to the original image (Misra and van der Maaten 2019). Predict relative position was another related method (C. Doersch 2015), and image rotation (S. Gidaris 2018) are demonstrated in (Figure 6).

\begin{figure}
  \centering
  \fbox{\includegraphics[scale=0.35]{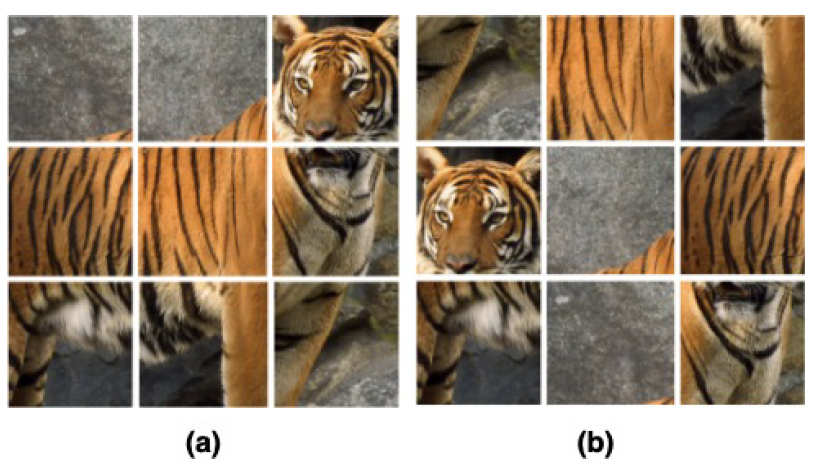}}
  \caption{One example of a Jigsaw puzzle.}
\end{figure}

\begin{figure}
  \centering
  \fbox{\includegraphics[scale=0.4]{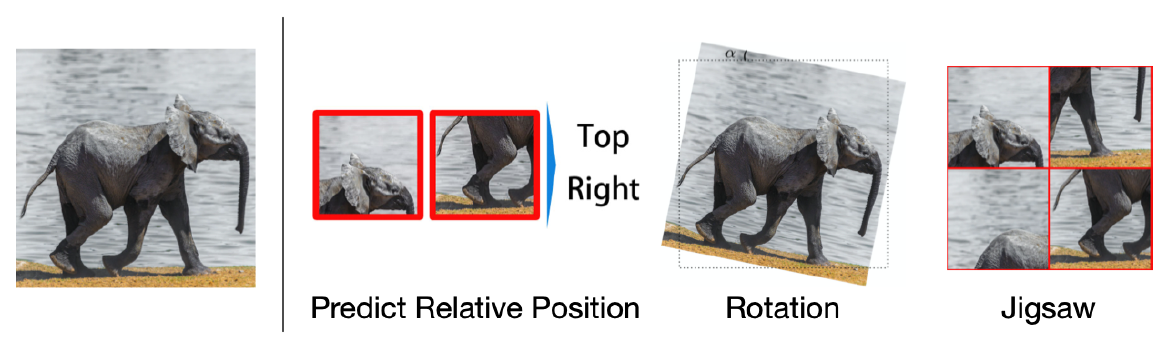}}
  \caption{Three main methods for spatial contrast: predict relative position, rotation, and Jigsaw.}
\end{figure}

\paragraph{Frame Order Based}
The frame ordered-based contrast is applied to time-varying data. These applications could be a sequence of sensor data (e.g., real-time MRI) or a picture frame (e.g. movie). Sensor data or a video with a sequence of linked frames that is closed on time stamp is more likely to be relevant than those that are distanced. This method is useful for solving a pretext task to learn the visual representation of a video or sensor data clip, which can then be used to restore the coherence of a video or real-time MRI clip.

Similar to the Jigsaw Puzzle section, we reshuffled the frame order from the original sequence of the image frames as the positive sample, and all the remaining frames in the dataset as the negative sample.

Another approach is to randomly sample two clips of the same length from the same video. The purpose is to train the model using the contrastive loss. If two clips from the same video are compared, they will be closer than clips from different videos (negative pairs). Qian's team studied a method for comparing the similarity of two positive samples to negative ones. In the embedding space, the contrastive loss is utilized to train the network to attract clips from the same video and repel clips from other videos (Qian, Meng et al. 2020):

$$l_{i,j}=-log \frac{exp(sim(\hat{z_i}, \hat{z_j})/\tau)}{\sum_{k=1}^{2N}1_{[k\neq i]} exp(sim(\hat{z_i}, \hat{z_k})/\tau)} $$

where $z_i$,$z_i'$ are the encoded representation of the two augmentations of clips of the i-th input video. The InfoNCE contrastive loss is denoted as $L=1/N \sum_{i=1}^N L_i sim(u,v)=(u^T v)/(||u||_2 ||v||_2 )$ is the inner product between normalized vectors. The positive pairs are two augmentations of the clips from the same frames (video).

The authors sample a temporal interval from a monotonically declining distribution from a raw video. The temporal interval is the number of frames between the start positions of two clips, and this interval is used to sample two segments from a video. The clips are then subjected to a temporally consistent spatial augmentation before being fed into a 3D backbone with an MLP head (Qian, Meng et al. 2020).

\begin{figure}
  \centering
  \fbox{\includegraphics[scale=0.3]{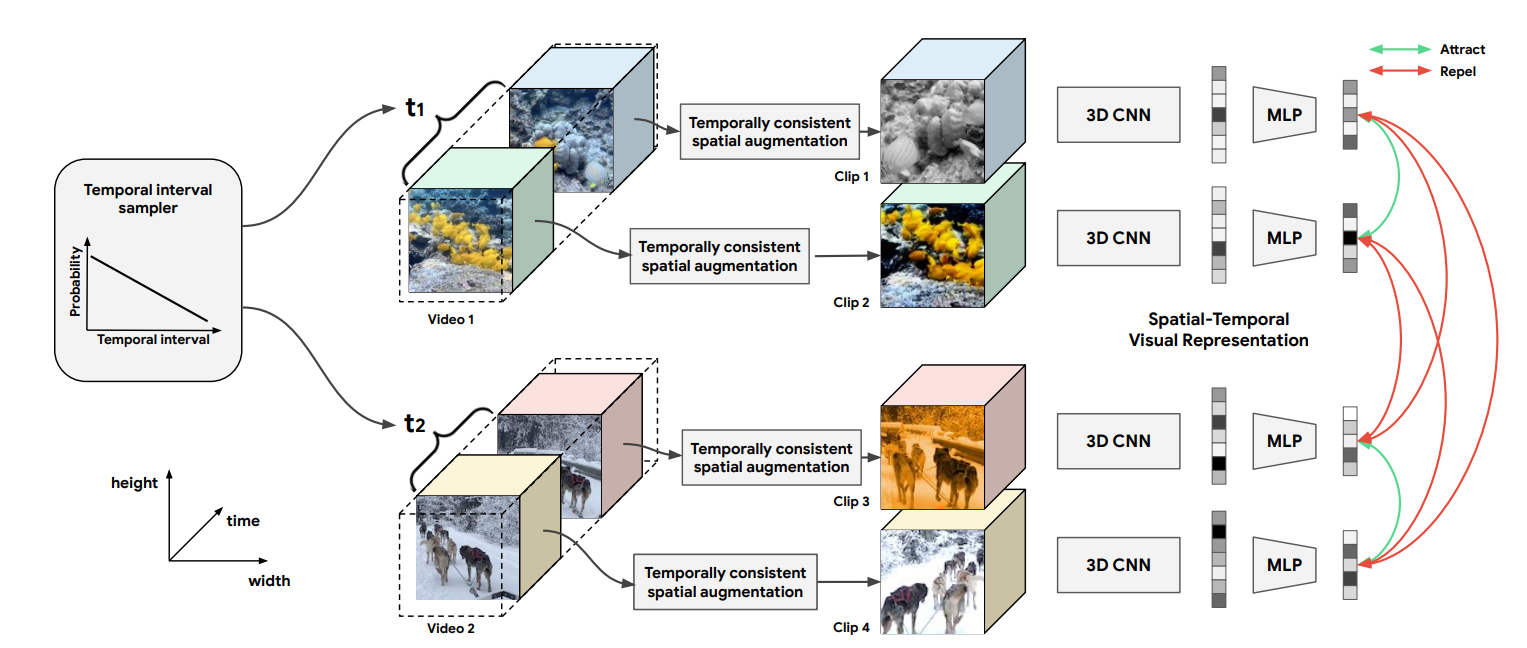}}
  \caption{Overview of the proposed spatiotemporal Contrastive Video Representation Learning (CVRL) framework.}
\end{figure}

\paragraph{Future Prediction}

\begin{figure}
  \centering
  \fbox{\includegraphics[scale=0.35]{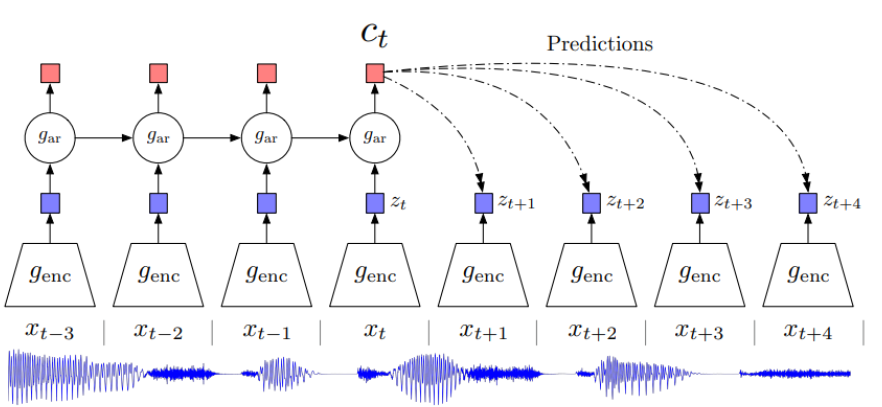}}
  \caption{Contrastive Predictive Coding.}
\end{figure}
Future prediction is the most often used strategy for evaluating future or missing information in data that varies over time. This method is frequently used with sequential data, such as sensor data, video, and so on. The task's purpose is to estimate information for the next time-stamp given a series of previous ones. One study creates a strategy for compressing high-dimensional data to low-dimensional latent space (van den Oord, Li et al. 2018). To summarize the information from the latent embedding space, auto-regressive models were used. The latent representation $C t$ is then created, as seen in **Figure 8**.During the task of predicting future information, the future information and $C t$ are encoded into a compressed vector representation that keeps the mutual information separate from the original information. Although **Figure 8** shows a strategy for dealing with an audio signal, same method may also be used for dealing with movies, photos, or text, etc. (van den Oord, Li et al. 2018).

\subsubsection{View Prediction}

\begin{figure}
	\centering
	\fbox{\includegraphics[width=0.48\textwidth]{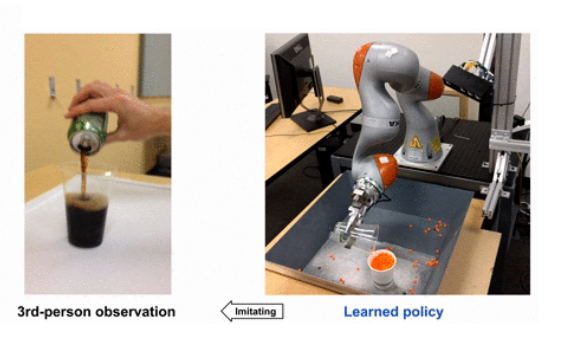}}
	\caption[width=0.2\linewidth]{The preview of results on imitation of human.}
\end{figure}

\begin{wrapfigure}{r}{0.5\textwidth}
	\centering
	\fbox{\includegraphics[width=0.48\textwidth]{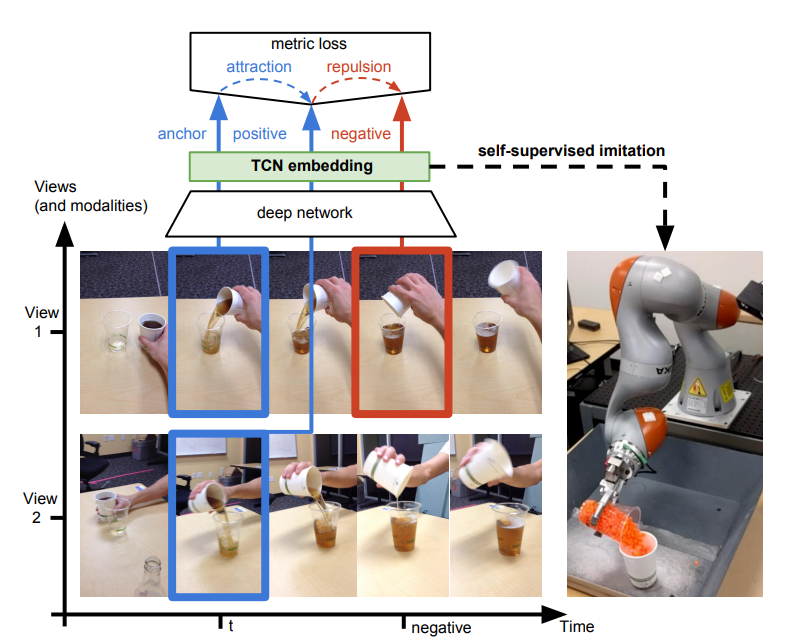}}
	\caption[width=0.2\linewidth]{Representation from video frames.}
\end{wrapfigure}

One recent study used a view prediction task on numerous views of the same video frame (Sermanet, Lynch et al. 2017). The purpose of this study is to imitate people by watching humans from a third-person perspective without human-annotation labels. The authors suggest such an algorithm for learning representation and robotic behavior from unlabeled films captured from several perspectives (1st and 3rd person) (Figure 9). To imitate human behavior, an invariant representation that captures the interactions between the robot's hands or grippers and the environment around them, the item, and human body signal was necessary.

Figure 10 shows the model's use of Time-Contrast Network (TCN) to learn the difference between similar-looking images in various time steps (red rectangle, Figure 10) and the difference between different-looking images at the same time frame (in the blue rectangle, Figure 10). This information allows their algorithm to detect traits that change over time rather than across distinct views.

The representation is learnt from unlabeled information in task-related movies, whereas robotic behaviors (e.g. pouring) are taught by viewing a single third-person viewpoint exhibited by a human. The authors present a reward function that enables reinforcement learning for robots to achieve the aim of the task, which is realistic in the real world, over the development of TCN.

\section*{References}

{
\small

[1] A. Krizhevsky, I. S., and G. E. Hinton (2012). "Imagenet classification with deep convolutional neural networks." NIPS: 1097–1105.

[2] C. Doersch, A. G., and A. A. Efros (2015). "Unsupervised visual representation learning by context prediction." In Proceedings of the IEEE ICCV: 1422–1430.

[3] Chen, T., et al. (2020). A Simple Framework for Contrastive Learning of Visual Representations. Proceedings of the 37th International Conference on Machine Learning. D. Hal, III and S. Aarti. Proceedings of Machine Learning Research, PMLR. 119: 1597--1607.

[4] Cimpoi, M. M., S.; Kokkinos, I.; Mohamed, S.; Vedaldi, A. (2014). "Describing textures in the wild." In Proceedings of the IEEE Conference on Computer Vision and Pattern Recognition.

[5] D. Mahajan, R. B. G., V. Ramanathan, K. He, M. Paluri, and A. B. Y. Li, and L. van der Maaten (2018). "Exploring the limits of weakly supervised pretraining." ECCV: 185–201.

[6] D. Pathak, P. K., J. Donahue, T. Darrell, and A. A. Efros, (2016). "Context encoders: Feature learning by inpainting." CVPR: 2536–2544.

[7] D. Tran, L. B., R. Fergus, L. Torresani, and M. Paluri, (2015). "Learning spatiotemporal features with 3D convolutional networks." ICCV.

[8] Epstein, R. (2016) The empty brain.

[9] Favaro, M. N. a. P. (2016). "Unsupervised learning of visual representions by solving jigsaw puzzles." ECCV.

[10] J. Long, E. S., and T. Darrell (2015). "Fully convolutional networks for semantic segmentation." CVPR: 3431–3440.

[11] J. S´anchez, F. P., T. Mensink, and J. Verbeek (2013). "Image classification with the fisher vector: Theory and practice." IJCV 105(3): 222–245.

[12] Liu, X., et al. (2020) Self-supervised Learning: Generative or Contrastive. arXiv:2006.08218

[13] Misra, I. and L. van der Maaten (2019) Self-Supervised Learning of Pretext-Invariant Representations. arXiv:1912.01991

[14] Noroozi, M. F., P. (2016). "Unsupervised learning of visual representations by solving jigsaw puzzles." Computer Vision—ECCV, Proceedings of the European Conference on Computer Vision.

[15] Qian, R., et al. (2020) Spatiotemporal Contrastive Video Representation Learning. arXiv:2008.03800

[16] R. Girshick, J. D., T. Darrell, and J. Malik (2014). "Rich feature hierarchies for accurate object detection and semantic segmentation." CVPR: 580–587.

[17] R. Zhang, P. I., and A. A. Efros (2016). "Colorful image colorization " ECCV: 649–666.

[18] S. Gidaris, P. S., and N. Komodakis (2018). "Unsupervised representation learning by predicting image rotations." ICLR.

[19] S. Gidaris, P. S., and N. Komodakis (2018). "Unsupervised representation learning by predicting image rotations." arXiv:1803.07728.

[20] Sermanet, P., et al. (2017) Time-Contrastive Networks: Self-Supervised Learning from Video. arXiv:1704.06888

[21] van den Oord, A., et al. (2018) Representation Learning with Contrastive Predictive Coding. arXiv:1807.03748

[22] W. Li, L. W., W. Li, E. Agustsson, and L. Van Gool (2017). "Webvision database: Visual learning and understanding from web data." arXiv:1708.02862.

}

\end{document}